\documentclass{ifacconf}

\usepackage{graphicx}      % include this line if your document contains figures
\usepackage{natbib}        % required for bibliography

\usepackage{algorithm} 
\usepackage{algorithmic}  
\usepackage[algo2e]{algorithm2e} 
\usepackage{graphicx}
\usepackage{framed}
\usepackage{makecell}
\usepackage{amsmath, amssymb}
\usepackage{newtxmath}
\usepackage{tabularx}
\usepackage{multicol}
\usepackage{comment}
\usepackage{romannum}
\usepackage{color}
\usepackage{enumitem}

\begin{document}
\begin{frontmatter}

\title{Prim-LAfD: A Framework to Learn and Adapt Primitive-Based Skills from Demonstrations for Insertion Tasks} 

% \thanks[footnoteinfo]{Sponsor and financial support acknowledgment goes here.}

\author[First]{Zheng Wu} 
\author[Second]{Wenzhao Lian} 
\author[First]{Changhao Wang}
\author[Third]{Mengxi Li} 
\author[Second]{Stefan Schaal} 
\author[First]{Masayoshi Tomizuka} 

\address[First]{Univerity of California, Berkeley,  
   Berkeley, CA 94709 USA (e-mail: \{zheng\_wu, changhaowang, tomizuka\}@berkeley.edu).}
\address[Second]{Intrinsic Innovation LLC, Mountain View, CA 94043 USA (e-mail: \{wenzhaol, sschaal\}@google.com)}
\address[Third]{Stanford University, Stanford, CA 94305  USA, (e-mail: mengxili@stanford.edu)}

\begin{abstract}                % Abstract of not more than 250 words.
Learning generalizable insertion skills in a data-efficient manner has long been a challenge in the robot learning  community. 
While the current state-of-the-art methods with reinforcement learning (RL) show promising performance in acquiring manipulation skills, the algorithms are data-hungry and hard to generalize.  
To overcome the issues, in this paper we present Prim-LAfD, a simple yet effective framework to learn and adapt primitive-based insertion skills from demonstrations. 
 Prim-LAfD utilizes black-box function optimization to learn and adapt the primitive parameters leveraging prior experiences. 
Human demonstrations are modeled as dense rewards guiding  parameter learning. We validate the effectiveness of the proposed method on eight peg-hole and connector-socket insertion tasks.
The experimental results show that our proposed framework takes less than one hour to acquire the insertion skills and as few as fifteen minutes to adapt to an unseen insertion task on a physical robot.
\end{abstract}

\begin{keyword}
Machine Learning, Learning for Control, Robotics, Motion Primitives, Learning from Demonstrations, Peg-in-Hole.
\end{keyword}

\end{frontmatter}
%===============================================================================

\section{Introduction}
\label{sec:intro}
	
Reinforcement learning (RL) has been widely used to acquire robotic manipulation skills in recent years~\cite{gu2017deep, kalashnikov2018scalable}. 
However, training an agent to perform certain tasks using RL is typically data-hungry and hard to generalize to novel tasks. 
This data-inefficiency problem significantly limits the adoption of RL on real robot systems, especially in real-world scenarios. 

Motion primitives, due to their flexibility and reliability, serve as a popular skill representation in practical applications~\cite{johannsmeier2019framework, Voigt2020learning, alt2021robot}. A motion primitive is characterized by a desired trajectory generator and an exit condition. It is often realized by hybrid motion/force controllers, such as a Cartesian space impedance controller. Taking moving until contact as an example primitive, the robot compliantly moves towards the surface until the sensed force exceeds a pre-defined threshold;
%A commonly used motion primitive is moving until contact where the robots compliantly moves towards a surface until the sensed force exceeds a threshold; 
a formal definition of motion primitives is deferred to Section~\ref{sec:method:primitive}. Despite the light-weighted representation and wide generalizability of the primitives, their parameters are often task-dependent, and the tuning requires domain expertise and significant trial-and-error efforts~\cite{lian2021benchmarking}.

In this work we present Prim-LAfD, a framework to learn and generalize insertion skills based on skill primitives. 
Recently, research efforts have been devoted to learning primitives for manipulation tasks~\cite{li2020learning, vuong2020learning}. However, most works treat primitives as a mid-level controller and use RL to learn the sequence of primitives. While these methods reduce the exploration space by leveraging primitives, they still suffer from the inherent drawbacks of RL: data inefficiency and lack of generalizability. %lots of training time and hard to generalize. 
To overcome data inefficiency issue, \cite{johannsmeier2019framework} proposed to use black-box optimizer to obtain the optimal primitive parameters in a pre-defined range by minimizing the task completion time. This approach achieves promising performance on the real robot and is shown to be more efficient than RL. 
% Moreover, since the optimization is framework is efficient and generalizable, it can be applied to different task setting robustly.
However, we empirically observe that the parameter range needs to be narrowly set; otherwise, the optimizer spends a long time exploring unpromising parameter choices. In addition, the objective (task completion time)  is a sparse reward, as it is only triggered when the task succeeds. This prevents the optimizer from extracting information from failed task execution trials, which requires a narrow parameter range to be carefully chosen.

\begin{figure*}
    \centering
    \includegraphics[width=.7\textwidth]{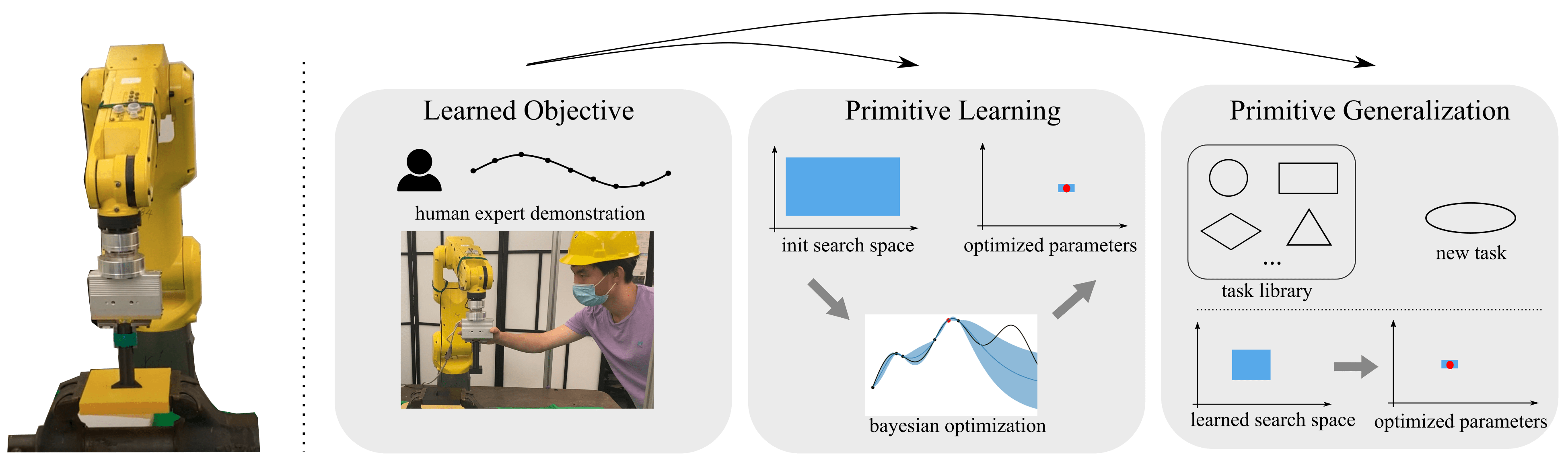}
    \caption{An overview of Prim-LAfD. Prim-LAfD learns a dense objective function from human demonstrations. It applies Bayesian Optimization (BO) to learn the primitive parameters w.r.t. the learned objective function. When generalizing to unseen tasks, we first select similar tasks from the task library based on the introduced similarity metric and then obtain a transferable search space for the new task for BO.}
    \label{fig:front}
    \vspace{-1mm}
\end{figure*}

Motivated from the above limitations, we propose a dense objective function that measures the likelihood of the induced execution trajectory sampled from the same distribution as the successful task demonstrations. 
% The intuition is to encourage the robot to follow the human expert demonstrations and further improve it via optimization.
This model-based objective function provides dense rewards even when a task execution trial fails, encouraging the optimizer to select parameters inducing execution trajectories more similar to the successful demonstrations, thus navigating the parameter space more efficiently. Furthermore, we propose a generalization method to adapt learned insertion skills to novel tasks via a task similarity metric, which alleviates the problem of requiring domain expertise to carefully set parameter ranges for novel tasks. In particular, socket (or hole) geometry information is extracted from the insertion tasks, and the $L_1$ distance between turning functions~\cite{arkin1991efficiently} of two hole shapes is used as the task similarity metric. An overview of the proposed Prim-LAfD is shown in Figure~\ref{fig:front}. Extensive experiments on $8$ different peg-hole and connector-socket insertion tasks are conducted to compare our proposed method with baselines. We experimentally demonstrate that Prim-LAfD can effectively and efficiently i) learn to acquire insertion skills with about $40$ iterations (less than an hour) training on a real robot and ii) generalize the learned insertion skills to unseen tasks with as few as $15$ iterations (less than $15$ minutes) on average.

% The contributions of our work are summarized as follows:
% \begin{enumerate}
%     \item We designed an insertion policy composed of learnable motion primitives leveraging impedance control. Rather than using sparse objective functions, we propose to learn a dense objective function from task demonstrations by modeling the execution trajectories.
%     \item We put forward a transfer learning method that retrieves similar tasks from a task library leveraging task meta information and adapts the previously learned parameters to the novel task with few interactions.
%     \item We collected 8 insertion tasks with diverse geometry shapes and clearance. Experimental results demonstrate that our method can acquire and adapt the insertion skills at practically acceptable time cost, $1$ hour and $15$ minutes respectively, while achieving a higher task success rate than baselines.
% \end{enumerate}

\section{Related Work}
\label{sec:related_work}
% \subsection{Learning Robotic Assembly Skills}
Robotic assembly tasks, e.g., peg insertion, have been studied for decades and are still one of the most popular and challenging topics in the robotics community~\cite{lian2021benchmarking}. 
Recently, many works have focused on developing learning algorithms for assembly tasks, among which deep reinforcement learning (RL) methods gained the most attention. 
For example,~\cite{lee2019making} proposes a self-supervised learning method to learn a neural representation from sensory input and used the learned representation as the input of deep RL.
% \cite{schoettler2019deep} combines a simple P-controller with a RL policy to reduce the exploration space of RL training. 
In~\cite{wu2020learning}, a reward learning approach is proposed to learn rewards from the high dimensional sensor input, and the reward is used to train a model-free RL policy. 
Nevertheless, these methods suffer from data inefficiency when facing novel tasks. There are also some works that combine RL and learning from demonstrations (LfD) to address the above issue~\cite{vecerik2017leveraging, davchev2020residual}. However, the impractical amount of robot-environment interactions required by deep RL algorithms and the domain expertise needed to adapt to different insertion tasks limit their adoption in real-world, particularly industrial scenarios.

By contrast, motion primitives, implemented with hybrid motion/force or impedance controllers, are often used for insertion tasks in practice. \cite{johannsmeier2019framework} makes the first attempt to learn the primitive parameters via black-box optimizers, minimizing the task completion time. This algorithm work effectively if primitive parameter ranges are narrowly set; otherwise, due to the sparse reward choice, it takes a large amount of robot-environment interactions for the optimizer to escape the parameter region leading to unsuccessful executions. In comparison, we propose a model-based dense objective function learned from demonstration, which guides the optimizer to explore more promising parameter regions earlier in training.
% Thus, building on the framework of \cite{johannsmeier2019framework}, we propose a learned dense objective function from demonstration, and develop a generalization method to avoid training from scratch for unseen tasks.

% \subsection{Generalizing Robotic Manipulation Skills to Unseen Tasks}
Transferring the existing policies to novel tasks is extensively studied in the field of robotic manipulation recently. It is especially important for RL based approaches, as most RL algorithms consider to learn a task in simulation first and then transfer the policy to the real world. One common approach is domain randomization, in which a variety of tasks are trained in simulation in order to capture the task distribution~\cite{tobin2017domain, peng2018sim}. Meta-RL has gained significant attraction in recent years~\cite{rakelly2019efficient}, where the experience within a family of tasks can be adapted to a new task in that family. However, for motion primitive learning methods, often optimized with gradient-free parameter search methods~\cite{johannsmeier2019framework, Voigt2020learning}, there haven't been efforts to transfer such prior experience to similar tasks. To the best of our knowledge, our work is the first attempt to encode the prior experiences and reduce the parameter exploration space during primitive learning on a novel insertion task.

\section{Proposed Method}
\label{sec:method}

In this section, we provide more details of  Prim-LAfD for insertion tasks. 
% Compared against traditional RL methods, our framework is more data-efficient and interpretable. 
% We deploy the proposed method on a wide range of peg-hole and connector-socket insertion (peg-in-hole for short) tasks for illustration, and it can be generalized to other manipulation tasks.
%though our framework can also be adopted to other manipulation tasks.

\subsection{Cartesian Impedance Control and the State-Action Space}

Impedance control is used to render the robot as a mass-spring-damping system following the dynamics below,
\begin{equation}
\label{eq:impedance}
   \mathbf{M} (\mathbf{\ddot{x}} - \mathbf{\ddot{x}}_d) + \mathbf{D} (\mathbf{\dot{x}} - \mathbf{\dot{x}}_d) + \mathbf{K} (\mathbf{x} - \mathbf{x_d}) = - \mathbf{F}_{ext},
\end{equation}
where $\mathbf{M}$, $\mathbf{D}$, $\mathbf{K} $ are the desired mass, damping, and stiffness matrices, and $\mathbf{F}_{ext}$ denotes the external wrench. $\mathbf{\ddot{x}}_d$, $\mathbf{\dot{x}}_d$, $\mathbf{x}_d$ are the desired Cartesian acceleration, velocity, and pose of the end-effector, and $\mathbf{\ddot{x}}$, $\mathbf{\dot{x}}$, $\mathbf{x}$ are the current values correspondingly. We assume a small velocity in our tasks and set $\mathbf{\ddot{x}}$, $\mathbf{\dot{x}}$ to 0, thus arriving at this control law,
\begin{equation}
\begin{split}
 \boldsymbol{\tau} & =  \mathbf{J}(\mathbf{q})^T \mathbf{F},\\
 \mathbf{F} & = -\mathbf{K} (\mathbf{x} - \mathbf{x_d}) - \mathbf{D} \mathbf{\dot{x}} + \mathbf{g} (\mathbf{q}),
 %+ \mathbf{B}(\mathbf{{q}} )  \mathbf{\ddot{q}} + \mathbf{C}(\mathbf{{q}}, \mathbf{\dot{q}}) \mathbf{\dot{q}} 
\end{split}
\end{equation}
where $ \boldsymbol{\tau}$ is the control torque, $\mathbf{F}$ is the control wrench, $\mathbf{J}(\mathbf{q})$ is the Jacobian, and $\mathbf{g} (\mathbf{q})$ is the gravity compensation force. 
% , where $\mathbf{B}(\mathbf{{x}} )$ is the mass-inertial matrix, $\mathbf{C}(\mathbf{{x}}, \mathbf{\dot{x}})$ is the Coriolis 

Throughout an insertion task, we would like to design a desired trajectory and a variable impedance to guide the robot movement. In favor of stability and ease of learning, we use a diagonal stiffness matrix $\mathbf{K}=\textbf{Diag}[K_x, K_y, K_z, K_{roll}, K_{pitch}, K_{yaw}]$, and, for simplicity, the damping matrix $\mathbf{D}$ is scaled such that the system is critically damped.

In summary, our insertion policy output, $\mathbf{a}_t\in\mathcal{A}$, fed to the impedance controller defined above, is composed of a desired end-effector pose $\mathbf{x}_d$ and the diagonal elements of the stiffness matrix $\mathbf{k} = \{K_x, K_y, K_z, K_{roll}, K_{pitch}, K_{yaw}\}$. The input to the policy, $\mathbf{s}_t\in\mathcal{S}$, consists of end-effector pose $\mathbf{x}_t$ and the sensed wrench $\mathbf{f}_t$, and is extensible to more modalities such as RGB and depth images.

\subsection{Manipulation Policy with Motion Primitives}
\label{sec:method:primitive}

In this section, we provide a detailed design on our insertion policy, which entails a state machine with state-dependent motion primitives. Each motion primitive $\mathcal{P}_m$ associated with the $m$-th state defines a desired trajectory $f_{\boldsymbol{\theta}_m}(\mathbf{x}_{enter}, \mathcal{T})$, an exit condition checker $h_{\boldsymbol{\theta}_m}(\cdot): \mathcal{S} \rightarrow \{1, 0\}$, and a 6-dimensional stiffness vector $\mathbf{k}_m$. $\boldsymbol{\theta}_m$ contains all the learnable parameters in the primitive $\mathcal{P}_m$. $\mathbf{x}_{enter}$ denotes the end-effector pose upon entering the $m$-th state. $\mathcal{T}$ contains the task information such as the 6 DoF poses of the peg and the hole; often, the hole pose defines the task frame of the motion primitives.

\begin{figure*}
    \centering
    \includegraphics[width=0.8\textwidth]{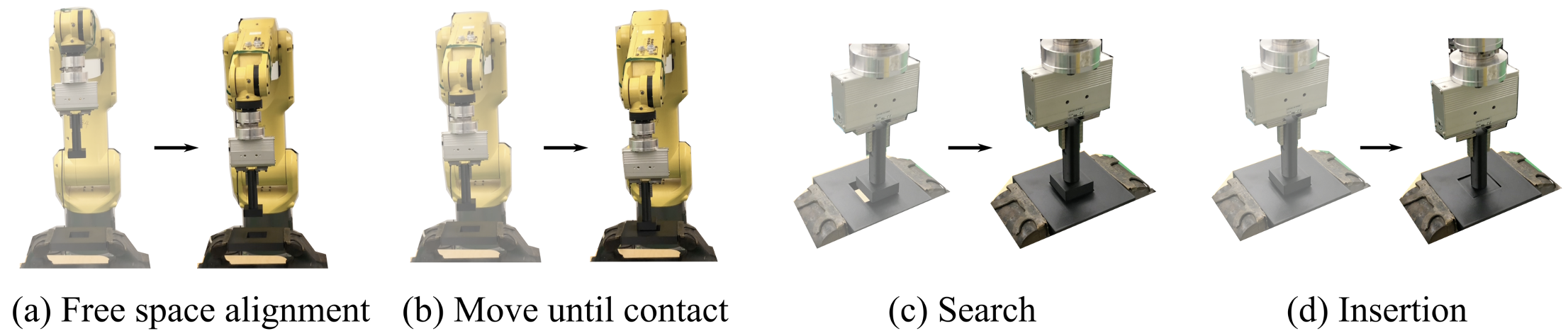}
    \caption{An illustrative figure of the motion primitives designed for peg-in-hole tasks. We show the start and the end states of the robot for each primitive.}
    \label{fig:primitive}
\vspace{-2mm}
\end{figure*}

In the following, we formally describe the $4$ motion primitives used in the peg-in-hole tasks, as shown in Figure~\ref{fig:primitive}.

\paragraph{Free space alignment.} The end-effector moves to an initial alignment pose.
\begin{equation}
\begin{split}
    f_{\boldsymbol{\theta}_1} &= u(\mathbf{x}_{enter}, \mathbf{x}_{target}) \\
    \quad  h_{\boldsymbol{\theta}_1} (\mathbf{s}_t) &= \mathbb{I} \left[ || \mathbf{x}_t - \mathbf{x}_{target} ||_2 < \sigma \right] \\
    \quad \mathbf{k}_1 &= \mathbf{k}_{max}.
\end{split}
\end{equation}
where $ \mathbb{I}[\cdot]$ is an indicator function mapping the evaluated condition to $\{0, 1\}$. $u(\cdot, \cdot)$ generates a linearly interpolated motion profile from the first to the second pose provided. The target end-effector pose $\mathbf{x}_{target}$ is extracted from the task information $\mathcal{T}$ as $\mathbf{x}_{target}=T_{hole}^{base} \cdot T^{hole}_{peg} \cdot T^{peg}_{ee}$, where $T_{hole}^{base}$ and $T^{peg}_{ee}$ denote the detected hole pose in robot base frame and the end-effector pose in peg frame. $ T^{hole}_{peg}$ is the desired peg pose in hole frame when the peg is above and coarsely aligned with the hole. $\mathbf{k}_{max}$ denotes a 6-dimensional vector composed of the highest stiffness values along each axis. $\sigma$ is a pre-defined threshold to determine if the robot arrives at the desired pose. No learnable parameters exist in this primtive. The parameters in this $1$-st primitive involves $\boldsymbol{\theta}_1=\{\O\}$.

\paragraph{Move until contact.} The end-effector moves towards the hole until the peg is in contact with the hole top surface.
\begin{equation}
\begin{split}
    f_{\boldsymbol{\theta}_2} &= u(\mathbf{x}_{enter}, \mathbf{x}_{enter} - \left[0 \; 0 \;  \delta \;  0 \;  0 \; 0\right]^T) \\
    h_{\boldsymbol{\theta}_2} (\mathbf{s}_t) &= \mathbb{I}[ f_{t, z} > \eta] \\
    \mathbf{k}_2 &= \mathbf{k}_{max}.
\end{split}
\end{equation}
$\delta$ is the desired displacement along z-axis in the task frame, $f_{t, z}$ is the sensed force along z-axis at time $t$, and $\eta$ is the exit force threshold. Therefore the parameters defining this $2$-nd primitive consists of $\boldsymbol{\theta}_2=\{\delta, \eta\}$.

\paragraph{Search.} The robot searches for the location of the hole while keeping in contact with the hole until the peg and the hole are perfectly aligned. After empirical comparisons with alternatives, including the commonly used spiral search, we choose the Lissajous curve as the searching pattern, which gives the most reliable performance. While searching for the translation alignment, the peg simultaneously rotates along the z-axis to address the yaw orientation error. The roll and pitch orientation errors are expected to be corrected by the robot being compliant to the environment with the learned stiffness.
\begin{equation}
\begin{split}
    f_{\boldsymbol{\theta}_3}(t) &= \mathbf{x}_{enter} + \left[\begin{matrix} A sin(2\pi a\frac{n_{1}}{T}t) \\ B sin(2\pi b \frac{n_{1}}{T}t) \\ -\gamma \\ 0 \\ 0 \\ \varphi sin(2\pi \frac{n_{2}}{T}t) \end{matrix}\right] \\ h_{\boldsymbol{\theta}_3} (\mathbf{s}_t)  &= \mathbb{I}[{x}_{enter, z} - {x}_{t, z} > \zeta] \\
    \mathbf{k}_3 &= \mathbf{k}_{search},
\end{split}
\end{equation}
% \begin{equation}
% \begin{split}
%     f_{\boldsymbol{\theta}}(t, \mathbf{s}_t) &=  \left[\begin{matrix} v_x(t, \mathbf{s}_t) \\ v_y(t, \mathbf{s}_t) \\v_z(t, \mathbf{s}_t) \\ \omega_x(t, \mathbf{s}_t) \\ \omega_y(t, \mathbf{s}_t) \\ \omega_z(t, \mathbf{s}_t) \\ f_x(t, \mathbf{s}_t) \\ f_y(t, \mathbf{s}_t) \\ f_z(t, \mathbf{s}_t) \\ \tau_x(t, \mathbf{s}_t) \\
%     \tau_y(t, \mathbf{s}_t) \\
%     \tau_z(t, \mathbf{s}_t) \\\end{matrix}\right] \\ h_{\boldsymbol{\theta}} (\mathbf{s}_t)  &= \mathbb{I}[{x}_{0, z} - {x}_{t, z} > \eta] \\
%     \mathbf{k} &= \mathbf{k}_{search}
% \end{split}
% \end{equation}
where $a=7, b=6$ are the Lissajous numbers selected and $T$ is the cycle period in Lissajous search, $\varphi$ is the maximum tolerated yaw error of the estimated hole pose, set as 6 degree in our experiments. The learnable parameters of this primitive are $\boldsymbol{\theta}_3=\{A, B, \frac{n_1}{T}, \frac{n_2}{T}, \gamma, \zeta, \mathbf{k}_{search}\}$.

\paragraph{Insertion.} The peg is inserted into the hole in a compliant manner.
\begin{equation}
\begin{split}
    f_{\boldsymbol{\theta}_4} &= u(\mathbf{x}_{enter}, \mathbf{x}_{enter} - \left[ 0 \; 0 \; \lambda \; 0 \; 0 \; 0 \right]^T) \\
    h_{\boldsymbol{\theta}_3} &= \mathbb{I}[\textrm{success condition}] \\
    \mathbf{k}_4 &= \mathbf{k}_{insertion},
\end{split}
\end{equation}
where the success condition is provided by the task information $\mathcal{T}$, e.g., $|| \mathbf{x}_t - \mathbf{x}_{success} ||_2 < \epsilon$. The primitive parameters to learn are $\boldsymbol{\theta}_4=\{\lambda, \mathbf{k}_{insertion}\}$.

\subsection{Learning Primitive Parameters}
\label{sec:method:learning}
In this section, we illustrate how to learn the primitive parameters $\boldsymbol{\Theta}=\{\boldsymbol{\theta}_1, \boldsymbol{\theta}_2, \boldsymbol{\theta}_3, \boldsymbol{\theta}_4\}$. The core idea is using a black-box optimizer to optimize a task-relevant objective function $J(\cdot)$.
% Throughout this section, we drop the explicit dependency of $J$ on $\Theta$ to simplify notations.
While a similar idea has been explored in~\cite{johannsmeier2019framework}, the objective function used is simply the measured task execution time. The major drawback of this objective function is that the objective signal is sparse and can only be triggered when the task is successfully executed. This makes the optimizer challenging to find a feasible region initially, especially when the primitive parameter space is large. Motivated by this, 
we propose a dense objective function that measures the likelihood of the induced execution trajectory being sampled from the distribution of successful task demonstrations
$\mathcal{E}_{D}=\{ \boldsymbol{\xi_i} \} (i = 1, 2, ..., M)$. Assuming the trajectroies are Markovian, a trajectory rollout $\boldsymbol{\xi} = [\boldsymbol{x}_0, \boldsymbol{x}_1, ..., \boldsymbol{x}_{n-1}]$ is modeled as:

\begin{equation}
    % \begin{aligned}
         p(\boldsymbol{\xi}; \boldsymbol{\Theta}) = p(\boldsymbol{x}_0) \prod_{i=1}^{n-1} p(\boldsymbol{x}_i | \boldsymbol{x}_{i-1}).
    %  p(\boldsymbol{\xi}; \boldsymbol{\Theta}) &= p(\boldsymbol{x}_0) \prod_{i=1}^{n-1} p(\boldsymbol{x}_i | \boldsymbol{x}_0, ..., \boldsymbol{x}_{i-1}) \\
    %       &= p(\boldsymbol{x}_0) \prod_{i=1}^{n-1} p(\boldsymbol{x}_i | \boldsymbol{x}_{i-1}).
% \end{aligned}
\end{equation}
    %  p(\boldsymbol{\xi}; \boldsymbol{\Theta}) &= p(\boldsymbol{x}_0) \prod_{i=1}^{n-1} p(\boldsymbol{x}_i | \boldsymbol{x}_0, ..., \boldsymbol{x}_{i-1}) \\
     %   &= p(\boldsymbol{x}_0) \prod_{i=1}^{n-1} p(\boldsymbol{x}_i | \boldsymbol{x}_{i-1}).
In order to learn $p(\boldsymbol{x}_i | \boldsymbol{x}_{i-1})$ from demonstrations, we first use a Guassian Mixture Model (GMM) to model the joint probability as $ p(\left[\begin{matrix} \boldsymbol{x}_i \\ \boldsymbol{x}_{i-1} \end{matrix}\right]) = \sum_{j=1}^{K}{\phi_j \mathcal{N}(\boldsymbol{\mu}_j, \boldsymbol{\Sigma}_j)}$ , where $\sum_{j=1}^{K}{\phi_j} = 1$, and $K$ is the number of GMM clusters.

We further represent the Gaussian mean $\boldsymbol{\mu}_j$ and variance $\boldsymbol{\Sigma}_j$ as: 
$\boldsymbol{\mu}_{\mathsmaller{j}} = \begin{bmatrix} \boldsymbol{\mu}^{\mathsmaller{1}}_{\mathsmaller{j}} \\ \boldsymbol{\mu}^{\mathsmaller{2}}_{\mathsmaller{j}} \end{bmatrix}$, 
$\boldsymbol{\Sigma}_{\mathsmaller{j}} = \begin{bmatrix} \boldsymbol{\Sigma}^{\mathsmaller{11}}_{\mathsmaller{j}} &  \boldsymbol{\Sigma}^{\mathsmaller{12}}_{\mathsmaller{j}} \\ \boldsymbol{\Sigma}^{\mathsmaller{21}}_{\mathsmaller{j}} & \boldsymbol{\Sigma}^{\mathsmaller{22}}_{\mathsmaller{j}} \end{bmatrix}$. 
We can then derive the conditional probability $p(\boldsymbol{x}_i | \boldsymbol{x}_{i-1}) = \sum_{j=1}^{K}{\phi_j \mathcal{N}(\overline{\boldsymbol{\mu}}_j, \overline{\boldsymbol{\Sigma}}_j)}$, where
\begin{equation}
\begin{split}
    \overline{\boldsymbol{\mu}}_j &= \boldsymbol{\mu}^1_{j} + \boldsymbol{\Sigma}^{12}_{j} (\boldsymbol{\Sigma}^{22}_{j})^{-1}(\boldsymbol{x}_{i-1} - \boldsymbol{\mu}^2_{j}) \\
    \overline{\boldsymbol{\Sigma}}_j &= \boldsymbol{\Sigma}^{11}_{j} - \boldsymbol{\Sigma}^{12}_{j} (\boldsymbol{\Sigma}^{22}_{j})^{-1} \boldsymbol{\Sigma}^{21}_{j}.
\end{split}
\end{equation}

Then, the objective function is designed as $J(\boldsymbol{\xi}) = \log{p(\boldsymbol{\xi}; \boldsymbol{\Theta})} + B$, where the first term encourages exploring parameters inducing similar trajectories to the successful demonstration traces, and the second term $B$ denotes a sparse bonus reward if the task succeeds.
% After obtaining the analytical form of $J(\boldsymbol{\xi}) = \log{p(\boldsymbol{\xi}; \boldsymbol{\Theta})}$, 
We use black-box optimizers to solve $\boldsymbol{\Theta}^*=\underset{\boldsymbol{\Theta}}{\mathrm{argmax}} J(\boldsymbol{\Theta})$, and Bayesian Optimization (BO) is selected in our work. Expected Improvement (EI) is used as the acquisition function, and we run BO for $N$ iterations. The learned parameter $\boldsymbol{\Theta}^*$ that achieves maximum $J(\boldsymbol{\Theta})$ during $N$ training iterations is selected as the optimal primitive configuration. Note that BO can be seamlessly replaced by other black-box optimization methods and the optimizer choice is not the focus of this work.

\subsection{Task Generalization}
\label{sec:method:generalization}
In this section, we detail our method on how to leverage prior experience when adapting to a novel insertion task, in particular, how to adapt previously learned peg-in-hole policies to different tasks with unseen hole shapes. Our adaptation procedure is composed of two core steps: measuring task similarities and transferring similar task policies to the unseen shape.

\subsubsection{Measuring task similarity}
\label{sec:method:generalization:measure_similarity}
Given an insertion skill library, i.e., a set of learned peg insertion policies for different shapes, $\mathcal{M} = \{\pi_1({\boldsymbol{\Theta}_1}), \pi_2({\boldsymbol{\Theta}_2}), ..., \pi_n({\boldsymbol{\Theta}_n})\}$ and an unseen shape, our goal is to first identify which subset of the $n$ tasks are most relevant to the new task.
While there is a diverse range of auxiliary task information that can be used to measure task similarity, here, we define the task similarity as the similarity between the hole cross-section contours. This assumption is based on the intuition that similar hole shapes would induce similar policies for insertion. For example, the insertion policies for a square hole and a rectangle hole are likely to be similar, and the optimal policy for a round hole might still work for a hexadecagon hole.
The similarity between a shape pair is measured by the $L_1$ distance between the two shapes' turning functions~\cite{arkin1991efficiently}.

% \begin{figure}
%     \centering
%     \includegraphics[width=0.3\textwidth]{figures/turning.pdf}
%     \caption{Examples of two hole's turning functions. Our task similarity metric is defined as the $L_1$ distance between the hole's turning functions.}
%     \label{fig:turning_function}
%     \vspace{-1mm}
% \end{figure}

Turning functions are a commonly used representation in shape matching, which represents the angle between the counter-clockwise tangent and the x-axis as a function of the travel distance along a normalized polygonal contour. 
% Two example turning functions are shown in Figure~\ref{fig:turning_function}. 
After obtaining the shape distances of the unseen shape and each shape in the task library, we choose the top $L$ shapes that are closest to the unseen shape as similar shapes. The policies of the similar shapes are then used as input for transfer learning detailed below in Section~\ref{sec:method:generalization:generalization_unseen}.
% \begin{wrapfigure}{r}{0.5\textwidth}
% % \begin{minipage}{\columnwidth}
% \label{fig:3}
% \centering
% \includegraphics[width=.4\textwidth]{figs/turning.pdf}
% \caption{Empty}
% % \end{minipage}
% \end{wrapfigure}

% \wenzhao{@zheng, worthwhile adding a note on how round shapes are treated, approximated as a polygon? Also, is it proper to say a turning function of a shape, rather than ... of a polygon}. An illustrative figure is shown in Fig.~\ref{fig:turning_function} \zheng{an illustrative figure might be needed here}. 
% Given the turning functions of shape $A$ and $B$, $\omega_A$ and $\omega_B$, the similarity is measured by the distance,
% \begin{equation}
%     d_{A, B} = \int {|\omega_A (x) - \omega_B (x)| dx}.
% \end{equation}

\subsubsection{Adapting to unseen shapes}
\label{sec:method:generalization:generalization_unseen}

Given a novel task, our goal is to efficiently adapt the already learned insertion policies of the most similar shapes. We build upon BO with hyperparameter transfer~\cite{perrone2019learning}. Unlike many works framing BO transfer learning as a multi-task learning problem, we attempt to learn the \textbf{search space} of BO from similar task policies and apply it to learning for the new task. 

Specifically, let $\mathcal{T} = \{ T_1, T_2, ..., T_t \}$ denote the task set of different hole shapes we selected as described in Section~\ref{sec:method:generalization:measure_similarity}, and $\mathcal{F} = \{ J_1, J_2, ..., J_t \}$ denotes the corresponding objective functions for each task. All the objective functions are initially defined on a common search space $\mathcal{X} \subseteq \mathrm{R}^{|\boldsymbol{\Theta}|}$, and it's assumed that we already obtained the optimal policies for the $t$ tasks $\{\pi_1(\boldsymbol{\Theta}_1^\star), \pi_2(\boldsymbol{\Theta}_2^\star), ..., \pi_t(\boldsymbol{\Theta}_t^\star)\} (\boldsymbol{\Theta}_i^\star \in \mathcal{X})$. Given an unseen task $T_{t+1}$, we aim to learn a new search space $\overline{\mathcal{X}} \subseteq \mathcal{X}$ from the previous tasks to expedite the new task learning process. We define the new search space as $\overline{\mathcal{X}} = \{ \boldsymbol{\Theta} \in \mathrm{R}^{|\boldsymbol{\Theta}|} | \mathbf{l} \leq \boldsymbol{\Theta} \leq \mathbf{u} \}$, where $\mathbf{l}, \mathbf{u}$ are the lower and upper bounds. It was proved in~\cite{perrone2019learning} that the new search space can be obtained by solving the constrained optimization problem:
\begin{equation}
    \min_{\mathbf{l}, \mathbf{u}} {\frac{1}{2} || \mathbf{u} - \mathbf{l} ||_2^2} \textrm{ such that for } 1 \leq i \leq t, \mathbf{l} \leq \boldsymbol{\Theta}_i^\star \leq \mathbf{u}.
\end{equation}
The optimization problem has a closed-form solution:
\begin{equation}
    \mathbf{l}^\star = \min\{\boldsymbol{\Theta}_i^\star \}_{i=1}^t, \mathbf{u}^\star = \max\{\boldsymbol{\Theta}_i^\star \}_{i=1}^t.
\end{equation}
This new search space is then utilized for policy training of this unseen shape task, following the procedure described in Section~\ref{sec:method:learning}.

\section{Experimental Results}
\begin{figure}
    \centering
    \includegraphics[width=0.3\textwidth]{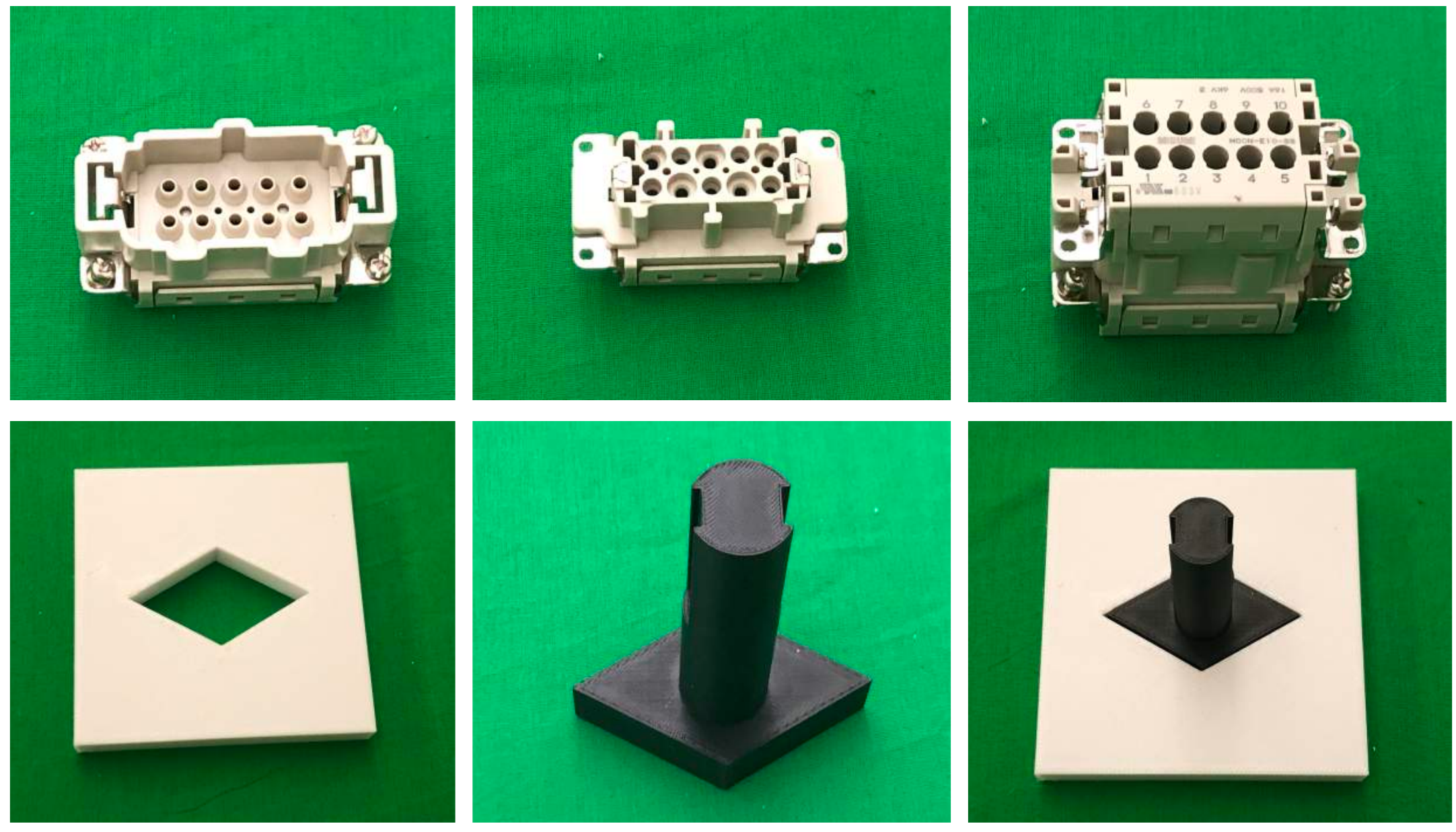}
    \caption{Two peg-hole pair instances (waterproof and parallelogram) used in our experiments.}
    \label{fig:exp_setup}
    \vspace{-1.mm}
\end{figure}

\
\label{sec:experiment}
We aim to investigate the effectiveness of Prim-LAfD by answering two questions: 1) whether the dense objective function proposed in Section~\ref{sec:method:learning} expedites the primitive learning process and improves policy performance, and 2) whether the
generalization algorithm described in Section~\ref{sec:method:generalization} is effective when transferring to an unseen shape. 
An insertion task library of 8 different peg-hole pairs is constructed, including 6 representative 3D-printed geometry shapes (round, triangle, parallelogram, rectangle, hexadecagon, ellipse) and two common industrial connectors (RJ45, waterproof). Examples are shown in Figure~\ref{fig:exp_setup}.

\subsection{Experimental Setup}
\label{sec:experiment:setup}
As shown in Figure~\ref{fig:front}, our hardware setup consists of a 6-DoF FANUC Mate 200iD robot and an ATI Mini45 Force/Torque sensor. The clearances for all 3D printed peg-hole pairs are 1mm; the waterproof and RJ45 are unaltered off-the-shelf connectors. To mimic the pose estimation error during industrial deployments, a uniform perturbation error of +/- 5mm in translation and +/- 6 degree in orientation is applied along each dimension.
The controller takes the policy output at 10 Hz and computes the torque command streamed to the robot at 1000Hz. All the learnable parameters of the policy and their initial range are listed in Table~\ref{table:param_list}.
Two metrics are used to evaluate the effectiveness and efficiency of the approaches: 1) the number of iterations the robot takes to accomplish the first successful insertion during training (denoted as \textit{number of iterations}), and 2) the success rate of the optimal policy after a fixed number of iterations (denoted as \textit{success rate}).

% \onecolumn
% \begin{table*}[!htbp]
\begin{table*}[hpbt]
\centering
\caption{Learnable parameters and corresponding range in the motion primitives.}
\label{table:param_list}
\scalebox{0.7}
{
\begin{tabularx}{1.25\textwidth}{c|c|c|c|c|c|c|c|c|c|c|c}
 &   \multicolumn{2}{c|}{\makecell{ \large Move until \\ \large contact}}     & \multicolumn{7}{c|}{ \large Search}  & \multicolumn{2}{l}{\large Insertion}                                      \\ \hline
\large Parameters   & \large $\delta \mathsmaller{(m)}$ & \large $\eta \mathsmaller{(N)}$ & \large $A\mathsmaller{(m)}$ & \large $B \mathsmaller{(m)}$ & \large $ n_1 / T \mathsmaller{(s^{-1})}$ & \large $n_2 / T \mathsmaller{(s^{-1})}$ & \large $\gamma \mathsmaller{(m)}$ & \large $\zeta \mathsmaller{(m)}$ & \large $\mathbf{k}_{search} \mathsmaller{(N/m, Nm/rad)}$ & \large $\lambda \mathsmaller{(m)}$ & \large $\mathbf{k}_{insertion}\mathsmaller{(N/m, Nm/rad)}$ \\ \hline
\large min   & \large $0$  & \large $1$ & \large $0$  & \large $0$  & \large $0 / 60$ & \large $0 / 60$  & \large $0$  & \large $0$  & \large $\mathbf{0}^{[6 \times 1]}$  & \large $0$  & \large $\mathbf{0}^{[6 \times 1]}$  \\ \hline
\large max & \large $0.1$ & \large $10$ & \large $0.02$ & \large $0.02$  & \large $2/10$ & \large $20/10$  & \large $0.02$ & \large $0.02$ & \large $[\mathbf{600}^{[3 \times 1]}, \mathbf{40} ^ {3 \times 1}]$  & \large $0.05$  & \large $[\mathbf{600}^{[3 \times 1]}, \mathbf{40} ^ {3 \times 1}]$   \\ \hline
\end{tabularx}
}
\end{table*}
% \twocolumn

\begin{figure*}[!htp]
    \centering
    \includegraphics[width=.65\textwidth]{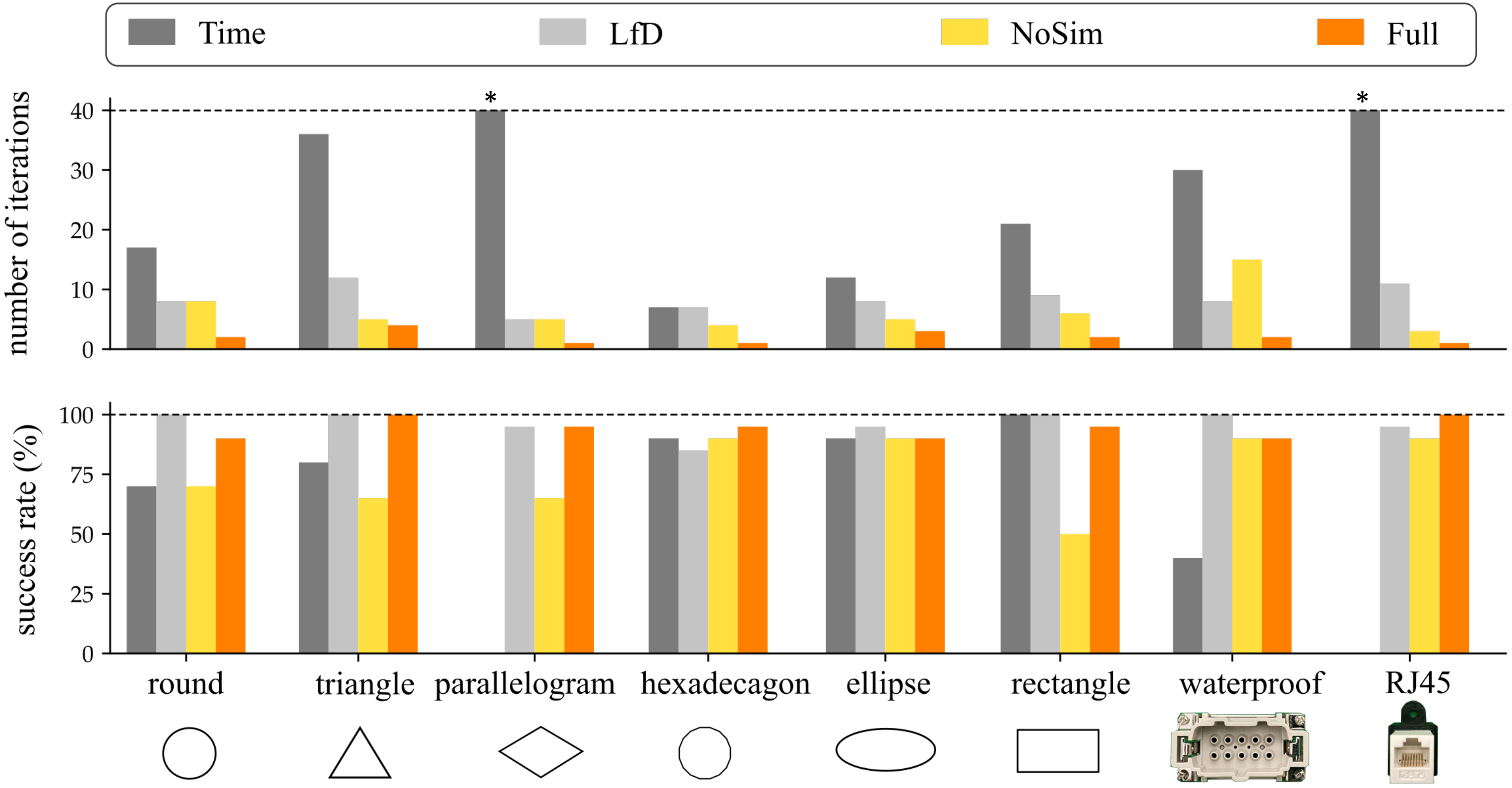}
    \caption{Experimental results for primitive learning and generalization. \texttt{Time}: primitive learning using task execution time as objective function, \texttt{LfD}: primitive learning using learned objective function (Section~\ref{sec:method:learning}), \texttt{NoSim}: primitive generalization without measuring task similarities, \texttt{Full}: full generalization method (Section~\ref{sec:method:generalization}). 
    $\star$ represents no successful trials is found during the learning process.}
    \label{fig:exp_results}
\vspace{-1.5mm}
\end{figure*}

\subsection{Learning Primitive Parameters}
\label{sec:experiment:learning}
In this experiment, we applied our learned objective function from demonstration (\texttt{LfD}) for primitive parameter optimization as described in Section~\ref{sec:method:learning}, and compare the results against primitive optimization by minimizing the measured task execution time (\texttt{Time})~\cite{johannsmeier2019framework}. When learning the objective function, we collected $10$ demonstrations for each insertion task through kinesthetic teaching. 
The demonstrations have $43.5$ time steps on average across different insertion tasks. 
The number of GMM clusters is set as $K=25$.

For each of the 8 peg-hole pairs, we run BO for 40 iterations. Within each iteration, the current policy is executed twice with independently sampled hole poses. The average objective of the two trials is used as the final objective value consumed by the BO step. The optimal policy is selected as the policy at the BO step achieving the optimal objective value, and evaluated by being executed 20 trials with independently sampled hole poses. As shown in Figure~\ref{fig:exp_results}, \texttt{LfD} outperforms \texttt{Time} on almost all the insertion tasks in both \textit{number of iterations} and \textit{success rate}. In some tasks, e.g., parallelogram and RJ45, \texttt{Time} cannot find proper parameters to achieve a single successful trial within 40 iterations, while \texttt{LfD} successfully navigates through the parameter space and accomplishes successful trials for all of the tasks. This validates the learned dense objective function, 
% i.e., a likelihood function measuring trajectory similarity w.r.t. the demonstrations, 
provides richer information for primitive learning than sparse signals alone like task success and completion time.

% Zheng: delete the content due to page budget

% \begin{figure}
%     \centering
%     \includegraphics[width=0.3\textwidth]{figures/learning_illustration.pdf}
%     \caption{Examples of two hole's turning functions. Our task similarity metric is defined as the $L_1$ distance between the hole's turning functions.}
%     \label{fig:learning_illustration}
%     \vspace{-3mm}
% \end{figure}

% To better illustrate the advantage of our learned objective function from demonstration, we show a sample trajectory during training in Figure~\ref{fig:learning_illustration}. The search trace thoroughly covers the the hole area and the corresponding parameters should be a reasonable candidate for the desired primitive configuration. However, the insertion did not succeed due to stochastic noise, and the optimizer discourages exploring the search space around this candidate in \texttt{Time}. By contrast, this candidate gets assigned a high objective value in \texttt{LfD} since its induced trajectory has a likelihood evaluated by our learned model from demonstration.

\subsection{Generalizing to Unseen Shapes}
% \wenzhao{The experimental results are lacking. At minimum, add a figure of task similarity, and a figure showing before and after learning, how the Lissajous search figure looks like, with sampled points overlayed. Please see the doc \hyperlink{https://docs.google.com/document/d/1m_0bTAGBC1giGotQo0o3tQFzOdNMC5xUBDj9BcaCTtA/edit?ts=60187449} for details.}
We now examine how our generalization method described in Section~\ref{sec:method:generalization} performs when transferring to unseen shapes. 
We consider the leave-one-out cross-validation setting, i.e., when presented 1 of the 8 tasks as the task of interest, the learning algorithm has access to all interaction data during policy learning on the other 7 tasks.
Two sets of experiments are conducted. 
First, we apply the full method described in Section~\ref{sec:method:generalization} (\texttt{Full}) to learn a reduced search space using the $L=3$ most similar tasks, within which the primitive parameters are optimized.
\label{sec:experiment:generalization}
% \begin{wrapfigure}{i}{0.4\textwidth}
% % \begin{minipage}{\columnwidth}
% \centering
% \includegraphics[width=.4\textwidth]{figs/confusion_matrix.pdf}
% \caption{The similarity distances between all insertion tasks.}
% \label{fig:confusion_matrix}
% % \end{minipage}
% \vspace{-4mm}
% \end{wrapfigure}

% \begin{figure}
%     \centering
%     \includegraphics[width=0.35\textwidth]{figures/confusion_matrix.pdf}
%     \caption{The similarity distances between all insertion tasks.}
%     \label{fig:confusion_matrix}
%     \vspace{-5mm}
% \end{figure}

Compared with \texttt{LfD} where parameters are optimized over the full space, \texttt{Full} reached a comparable or better success rate, meanwhile achieved the first task completion with a  lower number of iterations. 
Second, we consider learning the search space without measuring the task similarities (\texttt{NoSim}), i.e., the new search space is obtained using all the other tasks instead of only the similar ones.
% The purpose of this ablation study is to investigate the importance of task similarity in our generalization framework. 
% The results of the two set of experiments are shown in Table~\ref{table:primitive_generalization}. 
% Comparing \texttt{Full} with \texttt{Learn}, we can find that while both methods achieves similar results on the success rate of the optimal policy, \texttt{Full} requires much less iterations to acquire the first successful trial. It demonstrates our generalization method significantly speeds up the policy learning process. 
As seen in Figure~\ref{fig:exp_results}, \texttt{Full} outperforms \texttt{NoSim} consistently on all insertion tasks, indicating the significance of finding similar tasks using the task geometry information before learning the new search space. 
% The similarity distances between all insertion tasks are shown in Figure~\ref{fig:confusion_matrix} to qualitatively demonstrate the effectiveness of using the task geometry information. 
% For example, given an unseen waterproof insertion task, the triangle, rectangle, and parallelogram shapes are selected with the smallest similarity distances. 
% With such knowledge encoded, in \texttt{Full}, a narrower search space is obtained compared to \texttt{NoSim}, thus leading to better performance by focusing the BO budgets in the parameter subspace with a higher chance of task success.

\section{Conclusion}
\label{sec:conclusion}
% \paragraph{Summary.}
We propose Prim-LAfD, a data-efficient framework for learning and generalizing insertion skills with motion primitives. Extensive experiments on 8 different peg-hole and connector-socket insertion tasks are conducted to demonstrate the advantages of our method. The results show that Prim-LAfD enables a physical robot to learn peg-in-hole manipulation skills and to adapt the learned skills to unseen tasks at low time cost.

% \vspace{-2mm}
% \paragraph{Limitations and Future Work.}
% One limitation of the proposed generalization method is that the skill transfer is only beneficial across a narrow task family, e.g., peg-in-hole with different shapes. One future direction is generalizing the skills across more diverse tasks~\cite{lian2021benchmarking, suarez2016framework}. Besides, our current framework assumes an already designed primitive sequence composing the manipulation policy. Learning the primitive sequence from demonstration or self-supervisedly is another path to explore.

% \begin{ack}
% Place acknowledgments here.
% \end{ack}

\bibliography{ifacconf}             % bib file to produce the bibliography
                                                     % with bibtex (preferred)
                                                   
%\begin{thebibliography}{xx}  % you can also add the bibliography by hand

%\bibitem[Able(1956)]{Abl:56}
%B.C. Able.
%\newblock Nucleic acid content of microscope.
%\newblock \emph{Nature}, 135:\penalty0 7--9, 1956.

%\bibitem[Able et~al.(1954)Able, Tagg, and Rush]{AbTaRu:54}
%B.C. Able, R.A. Tagg, and M.~Rush.
%\newblock Enzyme-catalyzed cellular transanimations.
%\newblock In A.F. Round, editor, \emph{Advances in Enzymology}, volume~2, pages
%  125--247. Academic Press, New York, 3rd edition, 1954.

%\bibitem[Keohane(1958)]{Keo:58}
%\newblock \emph{Power and Interdependence: World Politics in Transitions}.
%\newblock Little, Brown \& Co., Boston, 1958.

%\bibitem[Powers(1985)]{Pow:85}
%\newblock Is there a way out?
%\newblock \emph{Harpers}, pages 35--47, June 1985.

%\bibitem[Soukhanov(1992)]{Heritage:92}
%\newblock \emph{{The American Heritage. Dictionary of the American Language}}.
%\newblock Houghton Mifflin Company, 1992.

%\end{thebibliography}

% \appendix
% \section{A summary of Latin grammar}    % Each appendix must have a short title.
% \section{Some Latin vocabulary}              % Sections and subsections are supported  
                                                                         % in the appendices.
\end{document}